\documentclass[11pt]{article}

\usepackage[preprint]{acl}
\usepackage{times}
\usepackage{latexsym}
\usepackage[T1]{fontenc}
\usepackage[utf8]{inputenc}
\usepackage{microtype}
\usepackage{graphicx}
\usepackage{multirow}
\usepackage{booktabs}
\usepackage{amsfonts}
\usepackage{amsmath}
\usepackage{xcolor}
\usepackage{float}
\usepackage{placeins}
\usepackage{tabularx}
\usepackage{array}
\usepackage{ragged2e}
\usepackage{makecell} 
\title{RAISE: RAG Design as an Architecture Search Problem}

\author{
\textbf{Zhen Chen\textsuperscript{1}\thanks{Equal contribution.}, Yibing Liu\textsuperscript{2}\footnotemark[1], Weihao Xie\textsuperscript{1}\footnotemark[1],}
\\[-1pt]
\textbf{Yu Liang\textsuperscript{2}, Peilin Chen\textsuperscript{1}, Shiqi Wang\textsuperscript{1}}\\[5pt]
\textsuperscript{1}City University of Hong Kong, Hong Kong SAR\quad \textsuperscript{2}Baidu Inc.\\[2pt]
\texttt{zchen979-c@my.cityu.edu.hk}
}

\begin{document}

\maketitle

\vspace{-10mm}

\begin{abstract}
Retrieval-augmented generation (RAG) systems expose numerous design choices spanning query rewriting, chunking, retrieval depth, reranking, and context compression. In practice, these choices are often configured through heuristics, hindering systematic evaluation and reproducibility across settings. We argue that this challenge is best formulated as RAG architecture search. To support controlled and reproducible study of this problem, we introduce the RAG Intelligence Search Engine (RAISE), a comprehensive framework and benchmark for RAG hyperparameter optimization, which evaluates optimization methods for RAG pipelines under standardized search spaces and budgets. RAISE implements 13 search algorithms and evaluates them across seven public text and multimodal datasets using three random seeds. Our experiments show that optimization performance is highly task-dependent: methods that perform strongly on one dataset may not generalize consistently across others, cautioning against interpreting aggregate rankings as evidence of universally superior strategies. RAISE provides a common experimental substrate for fair, reproducible, and systematic research on RAG hyperparameter optimization.
\end{abstract}

\section{Introduction}
\label{sec:introduction}

\begin{figure*}[!t]
    \centering
    \IfFileExists{figures/RAISE.pdf}{%
      \includegraphics[width=\textwidth, page=1]{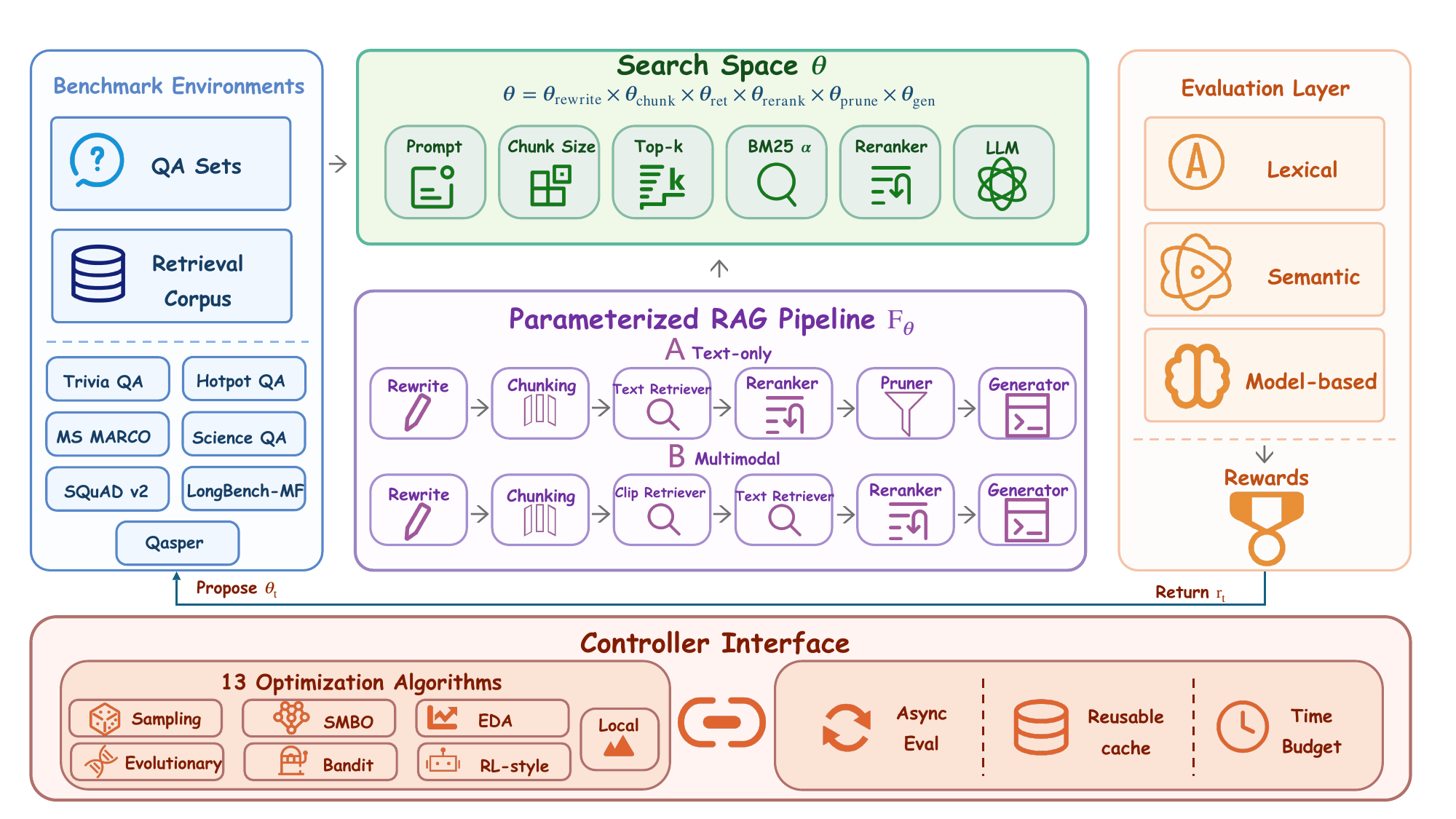}%
    }{%
      \fbox{\parbox{\linewidth}{\centering Missing figure file: \texttt{figures/RAISE.pdf}}}%
    }
    \caption{Overview of the RAG Intelligence Search Engine (RAISE). The framework couples a parameterized RAG pipeline, an evaluation layer that maps configurations to task-level rewards, and a controller interface for optimization algorithms.}
    \label{fig:ragsearch}
  \end{figure*}

Retrieval-augmented generation (RAG) grounds LLM outputs in external knowledge \cite{lewis2020rag}, but performance depends on tightly coupled choices such as chunker, query rewriting, retrieval depth, reranking, and context compression \cite{rag_hpo_impact}. These choices are often set through heuristics or trial-and-error tuning, making RAG optimization costly, difficult to reproduce, and challenging to compare across systems.

Recent work has explored RAG configuration through AutoRAG \cite{autorag} and hyperparameter impact analysis \cite{rag_hpo_impact}, adaptive retrieval through Self-RAG \cite{asai2024selfrag} and Adaptive-RAG \cite{jeong2024adaptive}, and automated tuning through AutoRAG-HP \cite{bandit_rag} and RAG-HPO analyses \cite{orbach2026raghpo}. However, these efforts use different pipelines, datasets, search spaces, optimization methods, and evaluation protocols, leaving the field without a shared framework for comparing optimization algorithms under controlled conditions. Such a framework should expose search spaces explicitly, standardize evaluation protocols, and support matched computational budgets. It should also remain extensible to emerging optimization algorithms and benchmark tasks, similar to standardized environments used in hyperparameter optimization research such as HPOBench \cite{eggensperger2021hpobench} and YAHPO Gym \cite{pfisterer2022yahpo}.

In this paper, we formulate RAG design as an architecture search problem. Instead of treating RAG tuning as an implementation detail, we formulate it as the selection of pipeline configurations that optimize end-to-end RAG performance under a fixed evaluation budget. This perspective connects RAG system design with random search \cite{bergstra2012random} and Bayesian optimization \cite{snoek2012practical}, while retaining challenges specific to RAG: heterogeneous modules, coupled design choices, multimodal pipelines, and task-dependent objective landscapes.

To enable systematic study of this problem, we introduce the RAG Intelligence Search Engine (RAISE), a comprehensive framework and benchmark for RAG hyperparameter optimization (Figure~\ref{fig:ragsearch}). RAISE makes the problem reproducible and comparable across algorithms and settings, while supporting both LLM and multimodal LLM pipelines. The framework is designed to remain extensible beyond the benchmark tasks included in this work. New optimization algorithms can be integrated through a common controller API, while new benchmark tasks can be added by specifying datasets, corpora, and evaluation protocols without changes to the core pipeline implementation. We benchmark 13 optimization algorithms on seven environments: TriviaQA \cite{joshi2017triviaqa}, HotpotQA \cite{yang2018hotpotqa}, MS MARCO \cite{bajaj2016msmarco}, ScienceQA \cite{lu2022learn}, SQuAD v2 \cite{rajpurkar2018know}, LongBench-Multifield \cite{bai2024longbench}, and LongBench-Qasper \cite{dasigi2021dataset}, with three seeds per setting.

We further conduct three-seed module ablations to identify which pipeline choices account for optimization gains under different task requirements. The results show that module effects vary with task structure: long-document retrieval benefits from query rewriting and pruning, while multi-hop reasoning relies more on retrieval depth. Likewise, long-context settings are especially sensitive to retrieval-depth control. These findings show that RAISE is not only a software framework but also a common experimental basis for studying RAG architecture search -- an open problem we hope will attract broader research attention.

In summary, our contributions are as follows:
\begin{itemize}
\setlength{\itemsep}{3pt}
\setlength{\parskip}{0pt}
\setlength{\parsep}{0pt}
    \item We establish RAG architecture search as a benchmark setting, enabling existing and future optimization algorithms to be studied on end-to-end RAG pipelines through RAISE's unified interface. The project code is available at \url{https://github.com/family99chen/RAISE}.
    
    \item We instantiate RAISE with a shared end-to-end RAG search space, public benchmark datasets, and a unified evaluation protocol. The benchmark spans query rewriting, chunking, retrieval, reranking, pruning, and generation, exposing heterogeneous task-specific optimization challenges under matched computational budgets.
    
    \item We conduct a controlled three-seed evaluation of 13 optimization algorithms across seven text and multimodal datasets, showing that optimizer performance is strongly environment-dependent. Our results motivate reporting RAG architecture search as optimizer--environment interactions rather than a universal leaderboard.
\end{itemize}

\section{Related Work}
\label{sec:related_work}

\paragraph{RAG architectures and adaptive retrieval.}
Retrieval-augmented generation (RAG) was introduced as a means of combining parametric language models with external retrieval \cite{lewis2020rag}. Subsequent work has shown that retrieval should often be adaptive rather than fixed: Self-RAG enables a model to decide when to retrieve and how to critique retrieved evidence during generation \cite{asai2024selfrag}, while Adaptive-RAG selects retrieval strategies based on question complexity \cite{jeong2024adaptive}. Such studies confirm that end-to-end RAG quality hinges on retrieval and control decisions. Their focus, however, lies in designing or learning particular RAG architectures, rather than in general optimization over configurable RAG pipelines.

\paragraph{RAG hyperparameter optimization and benchmark environments.}
Several recent studies have begun to investigate hyperparameter optimization for RAG. AutoRAG-HP formulates RAG hyperparameter tuning as an online multi-armed bandit problem and employs a hierarchical controller for online adaptation \cite{bandit_rag}. \citet{orbach2026raghpo} compare several hyperparameter optimization methods for RAG across multiple datasets and a large search space. More broadly, our formulation builds on black-box and hyperparameter optimization, including random search \cite{bergstra2012random}, Bayesian optimization \cite{snoek2012practical}, budget-aware bandit methods such as Hyperband \cite{li2018hyperband}, and hybrid approaches such as BOHB \cite{falkner2018bohb}. Benchmark suites such as HPOBench \cite{eggensperger2021hpobench} and YAHPO Gym \cite{pfisterer2022yahpo} further demonstrate the value of standardized environments for reproducible optimizer comparison. RAISE extends this benchmark perspective to end-to-end RAG pipelines, where coupled modules, expensive evaluations, and task-dependent objectives make controlled comparison especially important.

\paragraph{Benchmarking and evaluating RAG pipelines.}
A complementary line of work investigates how RAG systems should be evaluated. ARES decomposes RAG quality into context relevance, answer faithfulness, and answer relevance \cite{saadfalcon2024ares}, while RAGAs provides reference-free metrics for similar dimensions \cite{es2024ragas}. Benchmarks such as RGB \cite{chen2024rgb} and CRUD-RAG \cite{lyu2025crudrag} further expose failure modes and component sensitivities in retrieval-augmented systems. Collectively, these studies establish that RAG quality is multi-dimensional and cannot be adequately captured by a single QA score. Our goal is complementary: rather than benchmarking RAG models or evaluation metrics in isolation, we benchmark optimization methods over shared end-to-end RAG configuration spaces under fixed budgets and standardized protocols.

\section{Methodology}
\label{sec:framework}

We define RAG architecture search as optimizing a parameterized end-to-end RAG pipeline under a fixed evaluation protocol. RAISE realizes this view through three components: a pipeline abstraction that specifies the admissible configurations, an evaluation layer that maps configurations to task-level scores, and a controller interface through which optimization algorithms propose configurations and receive feedback from the environment. This separation lets us vary the search strategy while holding the pipeline family, benchmark data, and scoring rule fixed. We adopt the term \emph{optimization algorithm} for the search method itself and \emph{controller} for the same method once instantiated within RAISE.

\subsection{Search Space and Pipeline Abstraction}
\label{subsec:pipeline_components}

\begin{table}[!t]
\centering
\small
\setlength{\tabcolsep}{3pt}
\renewcommand{\arraystretch}{1.08}
\begin{tabularx}{\columnwidth}{@{}
>{\hsize=0.74\hsize\RaggedRight\arraybackslash}X
>{\hsize=0.84\hsize\RaggedRight\arraybackslash}X
>{\hsize=1.42\hsize\RaggedRight\arraybackslash}X
@{}} 
\toprule
Component & Hyperparameter & Values \\
\midrule
Rewriter & prompt template & $\{\text{P1}, \text{P2}, \text{P3}\}$ \\ \midrule
\multirow{2}[1]{*}{Chunker} & chunk size / & $\{256, 512, 1024, 2048\}$ / \\ 
 & chunk overlap & $\{0, 64, 128, 192\}$ \\ \midrule
\multirow{3}[3]{*}{Text Retriever}
 & \multirow{2}[0]{*}{model /} & \{all-MiniLM-L6-v2, all-MiniLM-L12-v2\} / \\
 & top-$k$ / & $\{1, 3, 5, 10, 20, 50\}$ / \\ 
 & BM25 weight $\alpha$ & $\{0.0, 0.25, 0.5, 0.75, 1.0\}$ \\ \midrule 
% \multirow{2}[2]{*}{Reranker} & \multirow{2}[2]{*}{model / top-$k$} & \{MiniLM-L6-v2, TinyBERT-L2-v2\} / $\{1, 3, 5, 10, 20, 50\}$ \\ \midrule
\multirow{2}[2]{*}{Reranker} & \multirow{2}[0]{*}{model /} & \{MiniLM-L6-v2, TinyBERT-L2-v2\} / \\ 
& top-$k$ & $\{1, 3, 5, 10, 20, 50\}$ \\ \midrule

\multirow{2}[0]{*}{LLM}  & \multirow{2}[0]{*}{LLM stack} & Qwen3 series for rewriting, pruning, and generation \\ \midrule
\multirow{2}[3]{*}{\makecell[l]{Multimodal\\LLM}} & \multirow{2}[3]{*}{\makecell[l]{Vision-LLM\\stack}}  & Qwen3-VL series + CLIP top-$k \in \{1, 3, 5, 10, 20, 50\}$ \\
% Multimodal LLM & Vision-LLM stack & Qwen3-VL series + CLIP top-$k \in \{1, 3, 5, 10, 20, 50\}$ \\
% LLM & LLM stack & Qwen3-4B-Instruct for rewriting, pruning, and generation \\ 
% Multimodal LLM & LLM + vision retrieval & Qwen3-VL-4B-Instruct + CLIP top-$k \in \{1, 3, 5, 10, 20, 50\}$ \\
\bottomrule
\end{tabularx}
\caption{Instantiated experimental search space, with shared dimensions listed once and LLM and multimodal LLM components listed.}
\label{tab:instantiated_search_space}
\end{table}

RAISE models LLM and multimodal LLM RAG workflows as modular directed acyclic graphs. Formally, the search space $\Theta$ is the Cartesian product of module-specific configuration spaces:
\begin{equation}
\label{eq:theta}
    \Theta = \Theta_{\text{rewrite}} \times \Theta_{\text{chunk}} \times \Theta_{\text{ret}} \times \Theta_{\text{rerank}} \times \Theta_{\text{prune}} \times \Theta_{\text{gen}}
\end{equation}

This factorization disentangles query reformulation, chunk granularity, retrieval depth and scoring, reranking, pruning, and generation. LLM pipelines follow Rewriter $\rightarrow$ Chunker $\rightarrow$ Text Retriever $\rightarrow$ Reranker $\rightarrow$ Pruner $\rightarrow$ Generator, while multimodal LLM pipelines incorporate CLIP retrieval and omit pruning to preserve cross-modal alignment. The text retriever combines sparse BM25 scoring \cite{robertson2009probabilistic} with dense sentence embeddings from Sentence-BERT \cite{reimers2019sentence} and MiniLM \cite{wang2020minilm}, and the multimodal branch uses CLIP-aligned text--image embeddings \cite{radford2021learning}. Detailed module definitions are provided in Appendix~\ref{app:module_defs}.

\subsection{Evaluation Objective}
\label{subsec:evaluation_module}

Given a dataset $\mathcal{D} = \{(q_i, Y^*_i)\}_{i=1}^{N}$ of queries and references, and a RAG pipeline parameterized by $\theta \in \Theta$, RAISE formulates the search problem via the following dataset-level objective:
\begin{equation}
\label{eq:objective}
    \theta^* = \arg\max_{\theta \in \Theta} \frac{1}{N} \sum_{i=1}^{N} \mathcal{E}\left( \mathcal{F}_{\theta}(q_i), Y^*_i \right)
\end{equation}
where $\mathcal{F}_{\theta}(q_i) = Y_i$ denotes the generated response and $\mathcal{E}$ the evaluation function. Crucially, the optimization target is the full RAG pipeline rather than any module considered in isolation.

RAISE supports lexical, semantic, model-based, and efficiency-oriented signals, allowing $\mathcal{E}$ to be chosen for the task at hand. Table~\ref{tab:instantiated_search_space} gives the search space used in our experiments. In the main benchmark, we use an equal-weight objective over ROUGE-L \cite{lin2004rouge}, METEOR \cite{banerjee2005meteor}, token-F1 \cite{rajpurkar2018know}, and BLEU \cite{papineni2002bleu}, as described in Section~\ref{subsec:main_results}. Full metric definitions are provided in Appendix~\ref{app:eval_metrics}.

\subsection{Benchmark Environments}
\label{subsec:benchmark_environments}

RAISE serves as both a configurable RAG pipeline and a benchmark environment for studying optimization behavior. Each environment is specified by a question--answer set and a retrieval corpus, keeping the interface simple while allowing tasks to differ in evidence structure, answer form, and modality. The suite covers six LLM QA tasks and one multimodal LLM QA task.

\paragraph{Benchmark construction.}
The benchmark is designed to expose bottlenecks in end-to-end RAG search rather than maximize dataset size. As shown in Table~\ref{tab:datasets}, the suite stresses long-document localization, multi-evidence composition, retrieval and reranking, abstention, long-context chunking and pruning, and visual grounding. The environments instantiate the seven datasets in Table~\ref{tab:datasets}.

\begin{table}[H]
\centering
\small
\setlength{\tabcolsep}{2pt}
\renewcommand{\arraystretch}{0.96}
\begin{tabularx}{\columnwidth}{@{}
>{\hsize=1.15\hsize\RaggedRight\arraybackslash}X
>{\hsize=0.45\hsize\Centering\arraybackslash}X
>{\hsize=0.70\hsize\Centering\arraybackslash}X
>{\hsize=1.00\hsize\RaggedRight\arraybackslash}X
>{\hsize=1.70\hsize\RaggedRight\arraybackslash}X
@{}}
\toprule
Dataset & Type & QA / Corpus & Task & Primary pressure point \\
\midrule
TriviaQA \cite{joshi2017triviaqa} & Text & 100 / 698 & Open-domain & Long documents and alias-rich answers \\
HotpotQA \cite{yang2018hotpotqa} & Text & 100 / 236 & Multi-hop & Multi-evidence composition \\
MS MARCO \cite{bajaj2016msmarco} & Text & 100 / 828 & Retrieval & Retrieval and reranking quality \\ 
ScienceQA \cite{lu2022learn} & Text-Vision & 100 / 100 & Science & Visual-textual grounding \\ 
SQuAD v2 \cite{rajpurkar2018know} & Text & 100 / 100 & Extractive & Hallucination control and abstention \\ 
LongBench-MF \cite{bai2024longbench} & Text & 100 / 100 & Long-context & Cross-field retrieval and pruning \\ 
LongBench-Qasper \cite{dasigi2021dataset} & Text & 100 / 100 & Scientific & Long-context reasoning and no-answer behavior \\ 
\bottomrule
\end{tabularx}
\caption{RAISE evaluation suite instantiated as proxy environments for end-to-end search.}
\label{tab:datasets}
\end{table}

\subsection{Supported Optimization Methods}
\label{subsec:supported_methods}

RAISE includes 13 preset optimization algorithms grouped into seven descriptive families. The labels in Table~\ref{tab:algo_taxonomy_13} reflect the search state and selection rule used by each method, rather than a single canonical HPO taxonomy. We do not aim to tune a single optimizer for RAG. Instead, we use a common interface to expose different search biases. Algorithm-level details are given in Appendix~\ref{app:controller_details}.

\subsection{Controller Interface and Execution}
\label{subsec:search_optimizer_interface}

Optimization algorithms use RAISE through a common controller interface. Let $\mathcal{A}$ denote a built-in or user-defined optimizer. Under this interface, $\mathcal{A}$ acts as a controller. At search iteration $t$, it proposes a complete pipeline configuration $\theta_t \in \Theta$, and the environment evaluates this configuration and returns a reward:
\begin{equation}
\label{eq:interface}
    r_t = \text{RAISE\_Env.evaluate}(\theta_t, \mathcal{D})
\end{equation}
Here $r_t \in \mathbb{R}$, or a vector-valued reward in multi-objective settings, is computed by the evaluation layer on dataset $\mathcal{D}$.

This interface casts end-to-end RAG configuration as a black-box optimization problem under the same search space, datasets, budgets, and reward definition for all methods. The execution engine supports the practical requirements of large-scale comparison, including asynchronous evaluation, reusable caches, bounded-time execution, and zero reward on failure, while keeping new controllers easy to add. Additional execution details are provided in Appendix~\ref{app:execution}.

\begin{table}[H]
\centering
\small
\setlength{\tabcolsep}{3pt}
\renewcommand{\arraystretch}{1.12}
\begin{tabularx}{\columnwidth}{@{}
>{\hsize=0.92\hsize\RaggedRight\arraybackslash}X
>{\hsize=0.74\hsize\RaggedRight\arraybackslash}X
>{\hsize=1.34\hsize\RaggedRight\arraybackslash}X
@{}}
\toprule
Algorithm & Family & Selection rule \\
\midrule
Random Search \cite{bergstra2012random} & Random sampling & Uniformly samples full configurations \\
\midrule
Greedy Search \cite{russell2010ai} & Local trajectory & Applies the best immediate local change \\
Coordinate Descent \cite{wright2015coordinate} & Local trajectory & Optimizes one dimension at a time \\
Simulated Annealing \cite{kirkpatrick1983annealing} & Local trajectory & Accepts downhill moves under a temperature schedule \\
Iterative Local Search \cite{lourenco2003iterated} & Local trajectory & Repeats local refinement from perturbed restarts \\
\midrule
TPE \cite{bergstra2011algorithms} & SMBO/Bayes. & Samples from densities fitted to good and bad regions \\
Cross-Entropy Method \cite{rubinstein1999crossentropy} & EDA-style & Updates a sampling distribution from elite configurations \\
Regularized Evolution \cite{real2019regularized} & Evolutionary & Mutates selected parents in an aging population \\
\midrule
Thompson Sampling \cite{thompson1933likelihood} & Bandit & Samples actions according to posterior uncertainty \\
UCB \cite{auer2002finite} & Bandit & Chooses actions by reward plus an exploration bonus \\
\midrule
GRPO \cite{shao2024deepseekmath} & RL-style & Updates a policy with group-relative advantages \\
Dr. GRPO \cite{liu2025drgrpo} & RL-style & Uses a more conservative GRPO-style update \\
Reinforce++ \cite{hu2025reinforcepp} & RL-style & Optimizes a stochastic policy with regularized policy gradients \\
\bottomrule
\end{tabularx}
\caption{Thirteen preset optimization algorithms in RAISE, grouped by family and selection rule.}
\label{tab:algo_taxonomy_13}
\end{table}

\section{Experiments}
\label{sec:experiments}

\subsection{Benchmark Protocol}
\label{subsec:exp_setup}
\label{subsec:datasets}
\label{subsec:optimizers}
\label{subsec:benchmark_protocol}

We use RAISE as a controlled benchmark for RAG architecture search. The experimental design follows system-level RAG automation work, including AutoRAG \cite{autorag}, AutoRAG-HP \cite{bandit_rag}, and RAG-HPO analysis \cite{orbach2026raghpo}, that treats the full pipeline as the optimization target. In contrast, RAISE fixes the controller interface, evaluation budget, and benchmark suite so that optimizers can be compared under matched conditions, rather than being used only to produce a single tuned pipeline. This design allows us to examine whether a shared search space and heterogeneous environments reveal optimizer--environment interactions that are difficult to identify in isolated tuning studies. If RAG architecture search is governed by task structure, then the best-performing strategy should vary across retrieval-centric, multi-hop, abstention-sensitive, multimodal, and long-context environments. We therefore organize the empirical study around four research questions: whether optimizer rankings depend on the environment (RQ1), whether the proxy construction provides a sufficiently stable signal (RQ2), whether explicit search improves over random sampling under matched budgets (RQ3), and which pipeline modules and search dimensions have the largest effect on performance (RQ4).

\begin{table*}[!t]
\centering
\small
\setlength{\tabcolsep}{0pt}
\renewcommand{\arraystretch}{1.14}
\begin{tabular*}{\textwidth}{@{\extracolsep{\fill}}llccccccccc@{}}
\toprule
Family & Algorithm &
Hotpot &
MSM &
SciQA &
SQuAD &
Trivia &
Qasper &
MF &
Wins & Rank \\
\midrule
Rand. & Random & 0.353{\tiny$\pm$0.043} & 0.232{\tiny$\pm$0.014} & \textbf{0.315{\tiny$\pm$0.007}} & 0.223{\tiny$\pm$0.009} & 0.401{\tiny$\pm$0.006} & 0.093{\tiny$\pm$0.005} & 0.336{\tiny$\pm$0.023} & 1 & 5.6 \\
\midrule
Local & Greedy & \textbf{0.417{\tiny$\pm$0.005}} & 0.199{\tiny$\pm$0.013} & 0.311{\tiny$\pm$0.006} & 0.223{\tiny$\pm$0.004} & \textbf{0.404{\tiny$\pm$0.016}} & 0.092{\tiny$\pm$0.005} & 0.339{\tiny$\pm$0.044} & 2 & 5.7 \\
Local & Coord. & 0.409{\tiny$\pm$0.020} & 0.231{\tiny$\pm$0.037} & 0.313{\tiny$\pm$0.004} & 0.224{\tiny$\pm$0.004} & 0.397{\tiny$\pm$0.023} & 0.094{\tiny$\pm$0.002} & 0.337{\tiny$\pm$0.036} & 0 & 4.6 \\
Local & SA & 0.368{\tiny$\pm$0.038} & 0.205{\tiny$\pm$0.017} & 0.266{\tiny$\pm$0.062} & 0.200{\tiny$\pm$0.017} & 0.402{\tiny$\pm$0.009} & 0.092{\tiny$\pm$0.007} & \textbf{0.362{\tiny$\pm$0.020}} & 1 & 6.9 \\
Local & ILS & 0.340{\tiny$\pm$0.070} & 0.207{\tiny$\pm$0.007} & 0.269{\tiny$\pm$0.054} & 0.204{\tiny$\pm$0.016} & 0.376{\tiny$\pm$0.024} & 0.084{\tiny$\pm$0.009} & 0.331{\tiny$\pm$0.028} & 0 & 10.6 \\
\midrule
SMBO & TPE & 0.359{\tiny$\pm$0.072} & 0.221{\tiny$\pm$0.003} & 0.269{\tiny$\pm$0.029} & 0.226{\tiny$\pm$0.004} & 0.387{\tiny$\pm$0.027} & 0.101{\tiny$\pm$0.007} & 0.362{\tiny$\pm$0.002} & 0 & 5.1 \\
EDA & CEM & 0.323{\tiny$\pm$0.014} & \textbf{0.239{\tiny$\pm$0.010}} & 0.230{\tiny$\pm$0.075} & 0.225{\tiny$\pm$0.002} & 0.390{\tiny$\pm$0.021} & 0.090{\tiny$\pm$0.008} & 0.348{\tiny$\pm$0.012} & 1 & 7.4 \\
\midrule
Evol. & Reg-Evo & 0.401{\tiny$\pm$0.020} & 0.218{\tiny$\pm$0.007} & 0.264{\tiny$\pm$0.034} & 0.224{\tiny$\pm$0.002} & 0.394{\tiny$\pm$0.009} & \textbf{0.102{\tiny$\pm$0.005}} & 0.359{\tiny$\pm$0.010} & 1 & 4.9 \\
\midrule
Bandit & TS & 0.350{\tiny$\pm$0.037} & 0.214{\tiny$\pm$0.012} & 0.255{\tiny$\pm$0.049} & 0.217{\tiny$\pm$0.006} & 0.376{\tiny$\pm$0.020} & 0.089{\tiny$\pm$0.006} & 0.345{\tiny$\pm$0.013} & 0 & 9.1 \\
Bandit & UCB & 0.307{\tiny$\pm$0.065} & 0.205{\tiny$\pm$0.023} & 0.248{\tiny$\pm$0.044} & 0.202{\tiny$\pm$0.012} & 0.389{\tiny$\pm$0.017} & 0.089{\tiny$\pm$0.008} & 0.334{\tiny$\pm$0.032} & 0 & 11.4 \\
\midrule
RL-style & GRPO & 0.333{\tiny$\pm$0.057} & 0.214{\tiny$\pm$0.027} & 0.288{\tiny$\pm$0.031} & \textbf{0.231{\tiny$\pm$0.005}} & 0.402{\tiny$\pm$0.017} & 0.086{\tiny$\pm$0.008} & 0.349{\tiny$\pm$0.011} & 1 & 6.3 \\
RL-style & Dr. GRPO & 0.380{\tiny$\pm$0.049} & 0.219{\tiny$\pm$0.030} & 0.276{\tiny$\pm$0.049} & 0.202{\tiny$\pm$0.008} & 0.395{\tiny$\pm$0.003} & 0.095{\tiny$\pm$0.006} & 0.348{\tiny$\pm$0.013} & 0 & 6.0 \\
RL-style & Reinforce++ & 0.372{\tiny$\pm$0.025} & 0.236{\tiny$\pm$0.010} & 0.252{\tiny$\pm$0.082} & 0.212{\tiny$\pm$0.014} & 0.394{\tiny$\pm$0.009} & 0.095{\tiny$\pm$0.005} & 0.328{\tiny$\pm$0.013} & 0 & 7.4 \\
\bottomrule
\end{tabular*}
\caption{Main benchmark results as an optimizer--environment interaction table. All methods use the same 30-evaluation budget within each environment. Scores are computed with the RAISE main objective, the equal-weight average of ROUGE-L, METEOR, token-F1, and BLEU. Cells report the mean over three seeds with standard deviation. Bold marks the within-environment winner. \textit{Wins} counts the number of dataset columns won by each method, and \textit{Rank} is the average within-dataset rank, where lower is better.}
\label{tab:main_benchmark_interaction}
\end{table*}

We instantiate the seven environments in Table~\ref{tab:datasets} as fixed proxy tasks from the corresponding public datasets, following HPOBench \cite{eggensperger2021hpobench} and YAHPO Gym \cite{pfisterer2022yahpo} benchmark practice. Each task contains 100 question--answer pairs and an associated retrieval corpus, yielding 700 QA instances and 2{,}162 corpus units in total. We evaluate the 13 optimization algorithms listed in Table~\ref{tab:algo_taxonomy_13} through a shared black-box interface, isolating optimizer behavior under matched conditions. Each algorithm receives 30 configuration evaluations under random seeds $11$, $22$, and $33$. The scalar objective gives equal weight to ROUGE-L, METEOR, token-F1, and BLEU. LLM environments use Qwen3-4B-Instruct \cite{qwen3technicalreport} for rewriting, pruning, and generation, while the multimodal LLM environment uses Qwen3-VL-4B-Instruct \cite{qwen3vltechnicalreport} with CLIP-based visual retrieval. Additional protocol details are provided in Appendix~\ref{app:benchmark_protocol_details}.

\subsection{Main Benchmark: Optimizer Environment Interactions (RQ1)} \label{subsec:main_results}

The main benchmark examines whether optimizer behavior in RAG hyperparameter optimization (RAG-HPO) is determined primarily by the optimization algorithm or by the structure of the RAG environment. To isolate this question, RAISE fixes the controller interface, evaluation budget, and scoring rule, while evaluating each optimizer on the corresponding text or multimodal search space for each environment. Under this protocol, differences across rows compare search behavior within the same environment, whereas differences across columns show how optimizer behavior changes across retrieval, reasoning, abstention, multimodal grounding, and long-context challenges.

Table~\ref{tab:main_benchmark_interaction} reports results on seven benchmark datasets: HotpotQA \cite{yang2018hotpotqa}, MS MARCO \cite{bajaj2016msmarco}, ScienceQA \cite{lu2022learn}, SQuADv2 \cite{rajpurkar2018know}, TriviaQA \cite{joshi2017triviaqa}, LongBench-Qasper \cite{dasigi2021dataset}, and LongBench-Multifield \cite{bai2024longbench}. All runs use a budget of 30 trials under random seeds $11$, $22$, and $33$. Each column corresponds to one environment and its primary challenge, and each row corresponds to an optimizer instantiated through the shared controller interface. Scores should be interpreted within each dataset column. The final two columns summarize the number of mean-score wins and the average within-dataset rank for each method. We therefore read the table as an optimizer--environment interaction map rather than as a single aggregate leaderboard. Detailed per-dataset seed results are provided in Appendix Tables~\ref{tab:hotpot_detailed_results}--\ref{tab:scienceqa_detailed_results}.

\paragraph{Results and analysis.} Table~\ref{tab:main_benchmark_interaction} shows that no optimizer dominates across environments. The best-performing method changes by task: Greedy Search leads on HotpotQA and TriviaQA, CEM on MS MARCO, Random Search on ScienceQA, GRPO on SQuADv2, Regularized Evolution on LongBench-Qasper, and Simulated Annealing on LongBench-Multifield. Under a fixed interface, budget, and scoring rule, these shifts indicate optimizer--environment interaction rather than a stable global ranking.

The aggregate columns support the same conclusion. Coordinate Descent achieves the best average rank without winning any individual dataset, while several other methods win only in specific settings. We therefore treat Table~\ref{tab:main_benchmark_interaction} as an interaction matrix that identifies where each search bias is useful.

\begin{figure*}[!t]
\centering
\includegraphics[width=\textwidth]{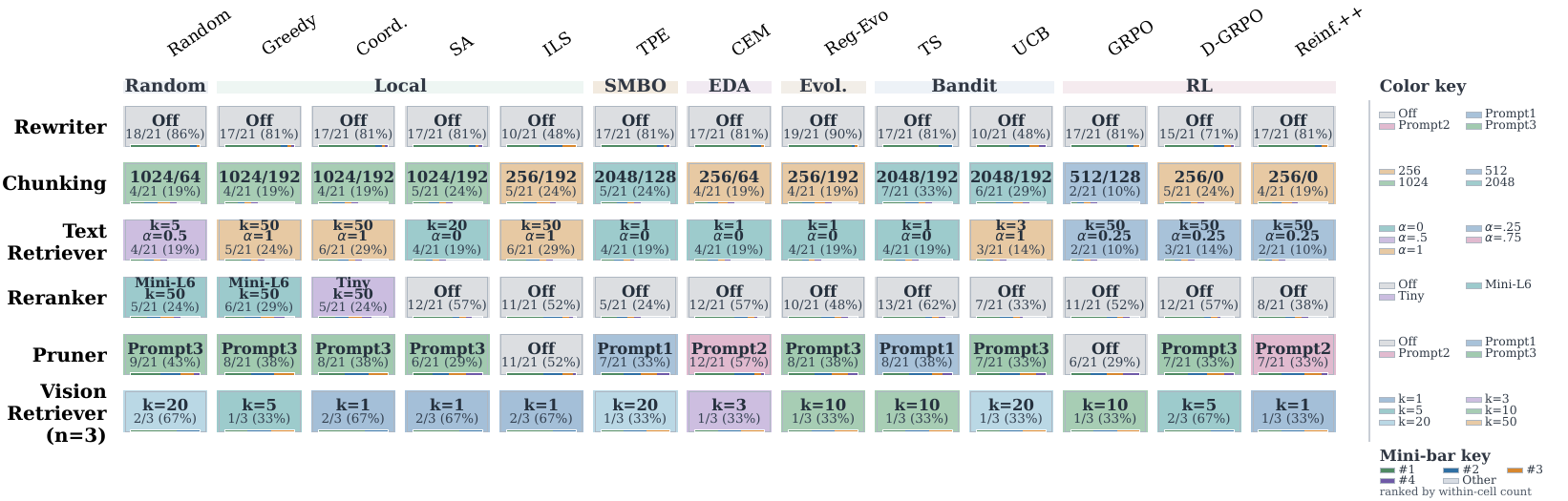}
\caption{Module-option choices in the final best configurations, aggregated over seven environments and three seeds ($N{=}21$; ScienceQA-only vision retriever $N{=}3$). Colors indicate option frequency and disabled settings; mini-bars show top alternatives. For retrieval and reranking, $k$ denotes top-$k$ candidates and $\alpha$ denotes the BM25 weight. Full distributions are reported in Appendix~\ref{app:module_preference_details}.}
\label{fig:module_option_preference_matrix}
\end{figure*}

Figure~\ref{fig:module_option_preference_matrix} shows that the optimizer--environment interaction in Table~\ref{tab:main_benchmark_interaction} also appears at the configuration level. Rewriting is frequently disabled, while retrieval, reranking, and pruning choices vary across methods and environments. Thus, comparable scores can arise from different pipeline choices, which motivates evaluating full configurations rather than isolated hyperparameters.

\begin{figure*}[!t]
\centering
\includegraphics[width=\textwidth]{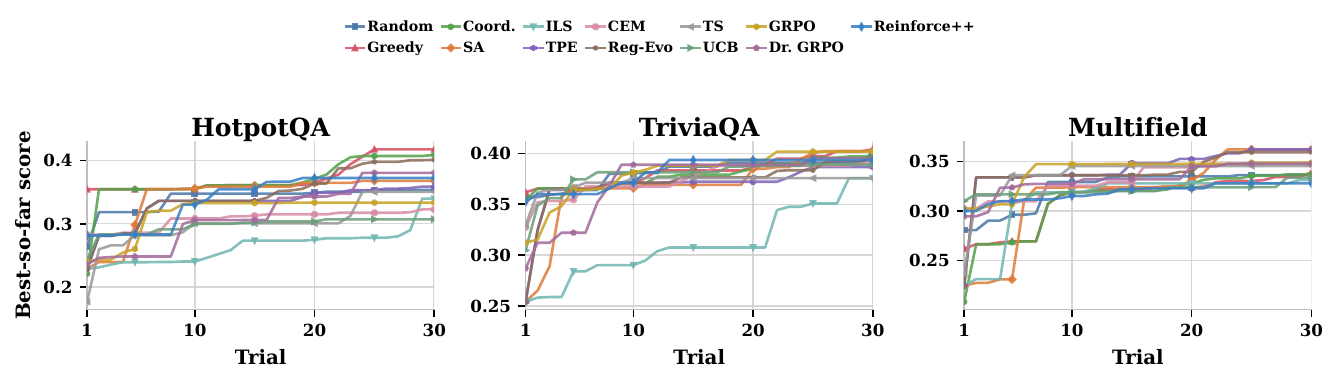}
\caption{Example best-so-far search trajectories under a 30-trial budget, showing how controllers improve as RAISE evaluates additional configurations.}
\label{fig:search_trajectory}
\end{figure*}

Figure~\ref{fig:search_trajectory} shows representative best-so-far trajectories over the same 30-trial budget.

\subsection{Ablation Study}
\label{subsec:ablation_experiments}

We complement the main benchmark with three ablation studies that examine the stability of proxy-task size and random seeds, the effect of explicit search relative to random sampling, and the contribution of individual pipeline modules to performance. Together, these analyses test whether the main findings are robust to proxy construction and reveal which parts of the RAG pipeline most strongly shape optimization outcomes.

\subsubsection{Robustness to QA Sample Size and Seeds (RQ2)}
\label{subsec:qa_size_stability}

To examine how proxy-task size affects optimization stability, we vary the HotpotQA subset size from 20 to 5{,}000 examples for CEM and TPE, using five random seeds and 30 trials per run. Figure~\ref{fig:qa_size_stability} shows substantial cross-seed variability for very small proxies, especially for TPE with 20 QA examples. At this scale, a small number of question instances can dominate the objective, making controller comparisons sensitive to seed selection. Stability improves at approximately 100--200 examples and then begins to saturate, while CEM remains more stable than TPE across the tested range. The flattening of the curves suggests that larger subsets mainly reduce noise once the proxy captures the main optimization signal. We therefore use the proxy as a controlled ranking screen rather than as a replacement for full-dataset evaluation. These results support the proxy protocol as a practical tradeoff between stability and cost for lightweight benchmarking, because it preserves the coarse optimizer signal without requiring full-corpus evaluation. This design keeps the benchmark lightweight while retaining enough task signal to compare optimizer behavior.

{\setlength{\intextsep}{5pt}
\begin{figure}[H]
\centering
\includegraphics[width=0.78\columnwidth]{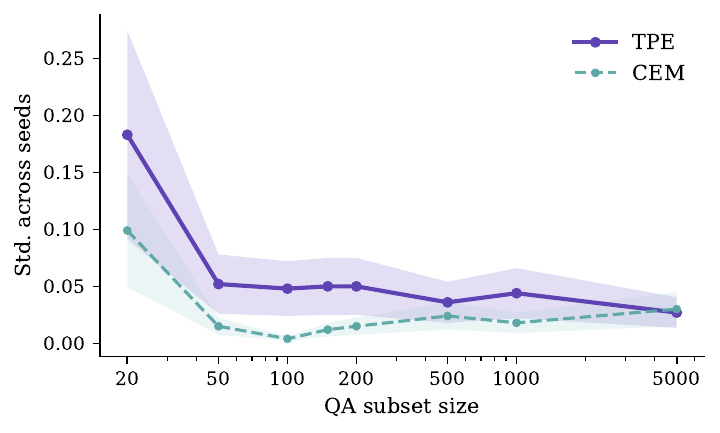}
\caption{Proxy-size stability across QA subset sizes.}
\label{fig:qa_size_stability}
\end{figure}
}

\subsubsection{Explicit vs. Random Search (RQ3)}
\label{subsec:random_average_ablation}

\begin{figure}[!htbp]
\centering
\includegraphics[width=0.94\columnwidth]{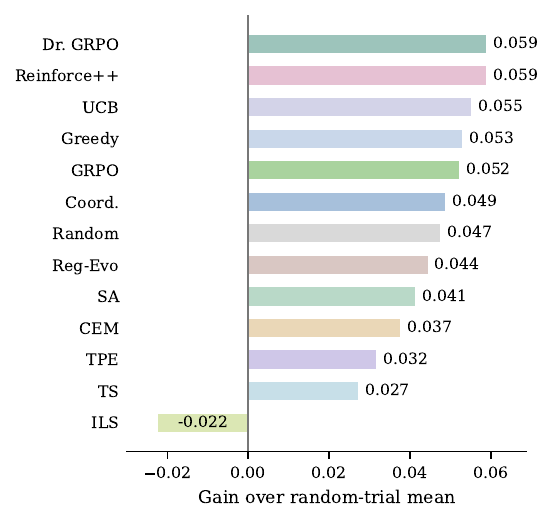}
\caption{Algorithm comparison over three random trials on the HotpotQA proxy task. Bars report gains over the random-trial mean.}
\label{fig:random_average_ablation}
\end{figure}

To test explicit optimization against unguided exploration, we compare search methods with random sampling on the HotpotQA proxy task under the same 20-evaluation budget and weighted objective. Figure~\ref{fig:random_average_ablation} shows that 11 of 12 non-random methods beat the baseline. The gains are modest but consistent, indicating that structured search can extract useful signal even when the budget is small. Adaptive methods tend to do better because they can spend later trials near more promising configurations. This advantage is budget-dependent rather than absolute. The result does not imply that random search is ineffective in general, but it shows that matched-budget comparisons should report random-trial baselines rather than relying only on optimizer rankings.

\subsubsection{Impact of Pipeline Modules (RQ4)}
\label{subsec:module_ablation}

\begin{figure}[t]
\centering
\includegraphics[width=0.94\columnwidth]{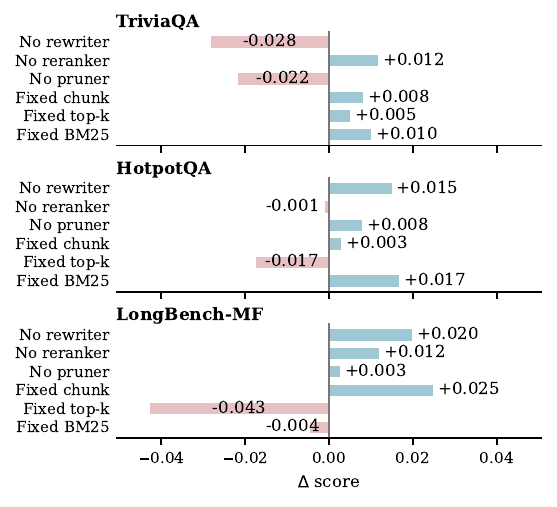}
\caption{Module sensitivity on TriviaQA, HotpotQA, and LongBench-Multifield. Bars show mean best-score changes relative to the full pipeline.}
\label{fig:module_ablation}
\end{figure}

We use module-level ablations on TriviaQA, HotpotQA, and LongBench-Multifield to examine how individual pipeline components affect performance under different task structures. Starting from the full text pipeline, we either remove the rewriter, reranker, or pruner, or fix one search dimension at a time. Each variant uses 13 optimization algorithms, three seeds, and 20 configuration evaluations.

Figure~\ref{fig:module_ablation} shows that module effects vary substantially across datasets. TriviaQA is most sensitive to rewriting and pruning, consistent with its long-document retrieval setting and alias-rich answers. HotpotQA is more sensitive to retrieval depth because it requires multi-hop evidence composition. LongBench-Multifield shows the largest drop when retrieval top-$k$ is fixed, whereas fixing chunk size is mildly beneficial. Across the three settings, no single module is uniformly dominant. Instead, the most influential search dimensions follow the evidence structure of the task, reinforcing the broader pattern that RAG pipeline choices are task-dependent. This finding cautions against treating any module choice as globally beneficial without specifying the environment and search budget used to evaluate it.

\section{Conclusion} \label{sec:conclusion}

In this paper, we have presented RAISE, a framework and benchmark that formulates RAG-HPO as black-box optimization over complete RAG pipelines. Across seven environments and 13 algorithms, we have shown that optimizer rankings vary with task structure and that no method is uniformly best. Our ablations have shown that proxy construction, random baselines, and module choices affect how gains are interpreted.

RAISE has turned RAG tuning into a repeatable benchmark problem. By separating the controller, search space, and environment, it has provided a shared testbed for comparing optimizers while reducing pipeline and evaluation confounds. This separation has made mixed results informative, because they show where search behavior transfers and where it remains environment-specific under fixed protocols. Accordingly, future RAG-HPO studies should report the search space, proxy construction, random baseline, and module constraints alongside final scores for interpretable, reproducible cross-study comparisons.

\clearpage

\section*{Limitations}
\label{sec:limitations}

This study has several limitations. Although RAISE is extensible, our experiments instantiate a fixed search space over specific RAG modules, models, and discrete values; alternative model families, retrieval backends, or continuous spaces may alter algorithm behavior. The main benchmark relies on lightweight proxy environments and a fixed budget, which enables broad comparison but does not substitute for full-dataset or larger-budget studies. Rankings are based on an equal-weight lexical and token-level objective, with semantic and model-based metrics reported as auxiliary signals. Finally, the variance analysis covers only selected settings and warrants extension across a wider range of datasets, budgets, and seeds.

\bibliography{reference}  % 导入你的 reference.bib 文件（注意：不要加 .bib 后缀）

%%%%%%%%%%%%%%%%%%%%%%%%%%%%%%%%%%%%%%%%%%%%%%%%%%%%%%%%%%%%

\appendix

\section{Module Preference Details}
\label{app:module_preference_details}

Figure~\ref{fig:module_option_preference_matrix} uses compact notation to keep the main-text matrix readable. In chunking cells, $a/b$ denotes chunk size $a$ and overlap $b$. In retrieval and reranking cells, $k=5$ denotes top-5 candidates, and $\alpha$ denotes the BM25 weight in the hybrid text retriever. P1--P3 denote the discrete prompt templates used by the rewriter or pruner. The stacked bars below expand the mini-bars in the main figure.

\begin{figure*}[t]
\centering
\includegraphics[width=0.98\textwidth]{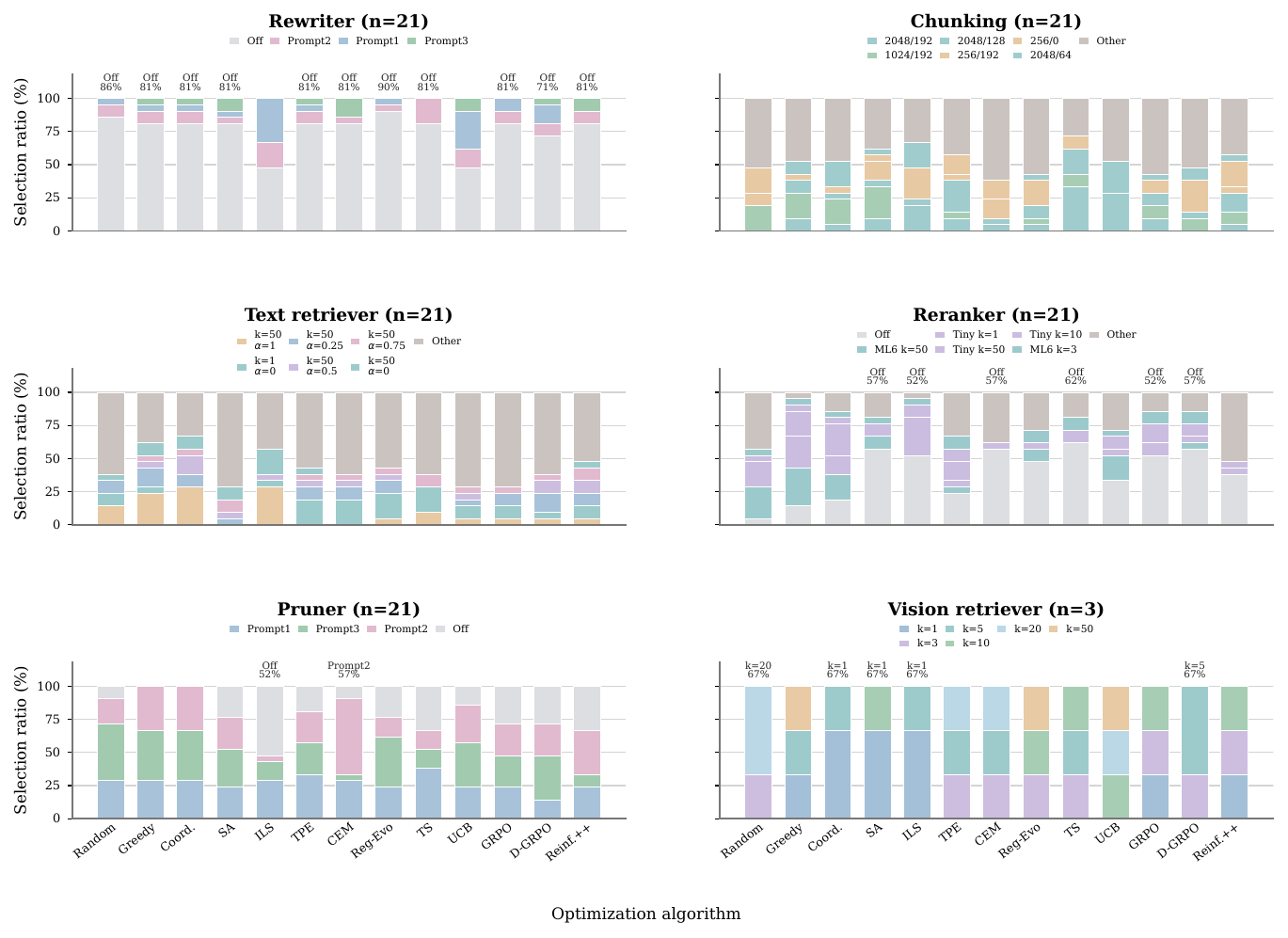}
\caption{Detailed module-option distributions behind Figure~\ref{fig:module_option_preference_matrix}. Each panel corresponds to one RAG module, each bar corresponds to one optimizer, and stacked segments show the normalized frequency of final options among the best configurations. Standard text modules aggregate seven environments and three seeds ($n=21$ per optimizer); the vision retriever is ScienceQA-only ($n=3$ per optimizer).}
\label{fig:module_preference_stacked}
\end{figure*}

\section{Detailed Framework Specification}
\label{app:framework_details}

\subsection{Pipeline Module Definitions}
\label{app:module_defs}

This appendix provides the detailed mathematical specification of the parameterized pipeline modules summarized in Section~\ref{sec:framework}.

\subsubsection{Prompt Template Search Space}
\label{app:prompt_templates}

RAISE treats prompt selection as a discrete search dimension for the rewriter and pruner. The generator prompt is fixed across text experiments, and the LLM-as-a-judge prompt is used only for auxiliary evaluation. Table~\ref{tab:prompt_template_search_space} lists the prompt templates exposed to the optimizer.

\begin{table*}[t]
\centering
\small
\setlength{\tabcolsep}{4pt}
\renewcommand{\arraystretch}{1.10}
\begin{tabularx}{\textwidth}{@{}
>{\hsize=0.45\hsize\RaggedRight\arraybackslash}X
>{\hsize=0.25\hsize\Centering\arraybackslash}X
>{\hsize=2.30\hsize\RaggedRight\arraybackslash}X
@{}}
\toprule
Module & ID & Prompt template \\
\midrule
Rewriter & P1 & Rewrite the user query for retrieval. Output only the rewritten query; do not answer the question or add explanations. \\
Rewriter & P2 & Rewrite the user query for retrieval with keywords and entities. Output only the rewritten query; do not answer the question or add explanations. \\
Rewriter & P3 & Rewrite the user query as a standalone question for retrieval. Output only the rewritten query; do not answer the question or add explanations. \\
\midrule
Pruner & P1 & Keep only sentences that directly support the answer. Output only the pruned context text; do not answer the question or add explanations. \\
Pruner & P2 & Select the minimal context needed to answer. Output only the pruned context text; do not answer the question or add explanations. \\
Pruner & P3 & Remove irrelevant content and keep key evidence only. Output only the pruned context text; do not answer the question or add explanations. \\
\midrule
Generator & fixed & Answer the question using only the provided context. If the answer is not in the context, state that it is unknown. \\
LLM judge & fixed & Judge whether the answer matches the reference list and return a JSON score with a short reason. \\
\bottomrule
\end{tabularx}
\caption{Prompt templates used by RAISE. Rewriter and pruner prompts are selected through the search space; generator and judge prompts are fixed for controlled evaluation.}
\label{tab:prompt_template_search_space}
\end{table*}

\subsubsection{Query Rewriter (Both Modalities)}
User queries in real-world scenarios are frequently ambiguous or underspecified. The Query Rewriter leverages LLMs to reformulate the user input prior to retrieval. Let $\mathcal{P} = \{p_1, p_2, \dots, p_N\}$ denote a discrete search space comprising $N$ candidate prompt strategies. Given an initial query $q$ and a selected prompt template $p_i \in \mathcal{P}$, the rewriter operates as a conditional generative function producing the optimized query $q'$:
\begin{equation}
\label{eq:rewrite}
    q' \sim \mathcal{M}_{\text{rewrite}}(\cdot \mid q, p_i)
\end{equation}
In RAISE, this categorical search space encompasses distinct reformulation paradigms, including standard retrieval reformulation, keyword or entity extraction, and standalone question generation.

\subsubsection{Chunker Module (Both Modalities)}
Text chunking critically affects retrieval granularity. Let $\mathcal{D}$ denote a document. The chunking function partitions $\mathcal{D}$ into a set of segments $\mathcal{C} = \{c_1, c_2, \dots, c_m\}$, parameterized by chunk size $s$ and overlap $o$:
\begin{equation}
\label{eq:chunk}
    \mathcal{C} = \text{Chunk}(\mathcal{D}; s, o)
\end{equation}
In our implementation, the search space includes $s \in \{256, 512, 1024, 2048\}$ tokens and $o \in \{0, 64, 128, 192\}$ tokens.

\subsubsection{Retrieval Stage: Text and Vision}
The retriever serves as the first-stage filter that narrows the candidate corpus to a manageable subset.

\paragraph{Text Retriever (Both Modalities).}
RAISE employs hybrid retrieval mechanisms that blend sparse and dense signals. For a rewritten query $q'$ and chunk $c \in \mathcal{C}$, the retrieval score is defined as
\begin{equation}
\label{eq:retrieval_score}
    s(q', c) = \alpha \cdot \text{BM25}(q', c) + (1-\alpha) \cdot \cos(E_q(q'), E_d(c))
\end{equation}
where $\alpha$ controls the interpolation between lexical and semantic retrieval. The initial retrieved set is then
\begin{equation}
\label{eq:text_topk}
    \mathcal{C}_{\text{ret}} = \mathop{\text{Top-}k_{\text{ret}}}_{c \in \mathcal{C}} s(q', c)
\end{equation}
with search parameters $\alpha \in \{0.0, 0.25, 0.5, 0.75, 1.0\}$ and $k_{\text{ret}} \in \{1, 3, 5, 10, 20, 50\}$.

\paragraph{Vision Retriever (Multimodal Only).}
For multimodal tasks, RAISE incorporates a vision-language retrieval stage based on CLIP-style embeddings. Given a visual corpus $\mathcal{V}$, the retrieved image subset is
\begin{equation}
\label{eq:vision_topk}
    \mathcal{V}_{\text{ret}} = \mathop{\text{Top-}k_{\text{vision}}}_{I \in \mathcal{V}} \cos(E_{\text{text}}(q'), E_{\text{vision}}(I))
\end{equation}
which provides visual evidence in parallel with text retrieval.

\subsubsection{Reranker (Both Modalities)}
To refine the first-stage retrieval output, the reranker applies a cross-encoder that jointly scores the query and candidate document \cite{nogueira2019passage}. Let $\mathcal{M}_{\text{cross}}$ denote the cross-encoder scoring function and $\oplus$ sequence concatenation. The reranked candidate set is
\begin{equation}
\label{eq:rerank}
    \mathcal{C}_{\text{rerank}} = \mathop{\text{Top-}k_{\text{rerank}}}_{c \in \mathcal{C}_{\text{ret}}} \mathcal{M}_{\text{cross}}(q' \oplus c)
\end{equation}
where the search space encompasses both the reranker choice and $k_{\text{rerank}} \in \{1, 3, 5, 10, 20, 50\}$.

\subsubsection{Pruner (Text Pipeline Only)}
Even after reranking, accumulated evidence may exceed the generator's context window or introduce distracting content. Let $\mathcal{P}_{\text{prune}} = \{p_1, p_2, \dots, p_M\}$ denote a discrete space of pruning prompt templates. The pruner applies a filtration function $f_{\text{prune}}$ to produce a condensed context:
\begin{equation}
\label{eq:prune}
    \mathcal{C}_{\text{final}} = f_{\text{prune}}(\mathcal{C}_{\text{rerank}}, q'; p_j)
\end{equation}
where $p_j \in \mathcal{P}_{\text{prune}}$ specifies the pruning strategy. The multimodal pipeline bypasses this stage to preserve visual--textual alignment.

\subsubsection{Generator}
The generator constitutes the final synthesis stage. Given the textual context $\mathcal{C}_{\text{final}}$, the retrieved visual context $\mathcal{V}_{\text{ret}}$ for multimodal tasks, and the original query $q$, generation follows the autoregressive factorization
\begin{equation}
\label{eq:generation}
    P(Y \mid q, \mathcal{C}_{\text{final}}, \mathcal{V}_{\text{ret}}) = \prod_{t=1}^{T} P_{\theta_{\text{gen}}}(y_t \mid y_{<t}, q, \mathcal{C}_{\text{final}}, \mathcal{V}_{\text{ret}})
\end{equation}
where $\mathcal{V}_{\text{ret}} = \emptyset$ in LLM pipelines.

\subsection{Evaluation Metrics}
\label{app:eval_metrics}

This appendix provides the metric definitions summarized in Section~\ref{subsec:evaluation_module}. Let $Y$ denote the predicted answer and $Y^*$ the reference answer or reference context, depending on the task.

\subsubsection{Lexical and N-gram Metrics}

\paragraph{Exact Match (EM).}
For extractive question answering, exact match is defined as
\begin{equation}
\label{eq:em}
    \text{EM}(Y, Y^*) = \mathbb{I}(Y = Y^*)
\end{equation}

\paragraph{Token-F1.}
Token-F1 treats $Y$ and $Y^*$ as bag-of-words and computes
\begin{equation}
\label{eq:f1}
    F_1(Y, Y^*) = \frac{2 \cdot |Y \cap Y^*|}{|Y| + |Y^*|}
\end{equation}

\paragraph{ROUGE-L.}
ROUGE-L \cite{lin2004rouge} is based on the longest common subsequence (LCS), with
\begin{equation}
\label{eq:rougel}
    F_{\text{LCS}} = \frac{(1+\beta^2) R_{\text{LCS}} P_{\text{LCS}}}{R_{\text{LCS}} + \beta^2 P_{\text{LCS}}}
\end{equation}
where $R_{\text{LCS}} = \frac{\text{LCS}(Y, Y^*)}{|Y^*|}$ and $P_{\text{LCS}} = \frac{\text{LCS}(Y, Y^*)}{|Y|}$.

\paragraph{BLEU.}
BLEU \cite{papineni2002bleu} is computed as
\begin{equation}
\label{eq:bleu}
    \text{BLEU} = \text{BP} \cdot \exp \left( \sum_{n=1}^{N} w_n \log p_n \right)
\end{equation}

\paragraph{METEOR.}
METEOR \cite{banerjee2005meteor} is defined as
\begin{equation}
\label{eq:meteor}
    \text{METEOR} = \left( \frac{10 P_m R_m}{R_m + 9 P_m} \right) (1 - \text{Pen})
\end{equation}

\subsubsection{Semantic and Model-Based Metrics}

\paragraph{BERTScore-F1.}
BERTScore \cite{zhang2020bertscore} computes semantic overlap using contextualized embeddings. Its recall term is
\begin{equation}
\label{eq:bertscore}
    R_{\text{BERT}} = \frac{1}{|Y^*|} \sum_{y^* \in Y^*} \max_{y \in Y} \mathbf{y}^\top \mathbf{y}^*
\end{equation}

\paragraph{LLM-as-a-Judge (LLM-aaJ).}
For complex reasoning tasks, RAISE supports an LLM-based judge model $\mathcal{M}$ whose output is parsed into a scalar reward:
\begin{equation}
\label{eq:llmjudge}
    \mathcal{E}_{\text{LLM}}(q, Y, R) = \Phi \left( \mathcal{M}(\text{prompt} \oplus q \oplus Y \oplus R) \right) \in \{0, 1\}
\end{equation}

\subsubsection{System and Search Efficiency}

\paragraph{End-to-End Search Time.}
The total wall-clock time of the search process is
\begin{equation}
\label{eq:searchtime_appendix}
    \mathcal{T}_{\text{total}} = \sum_{k=1}^{K} t_k + \mathcal{T}_{\text{overhead}}
\end{equation}

\paragraph{Configuration Evaluation Count.}
The total number of completed evaluations over $S$ search iterations is
\begin{equation}
\label{eq:evalcount_appendix}
    K = \sum_{s=1}^{S} \mathbb{I}(\text{eval}(\theta_s))
\end{equation}

\subsection{Execution Details and Robustness}
\label{app:execution}

Evaluating large numbers of configurations over full datasets incurs substantial computational overhead. To support large-scale search, the RAISE execution engine employs asynchronous concurrent execution to improve throughput, reusable caching to avoid redundant computation, and bounded-time evaluation to prevent individual failures from stalling the optimization loop.

Specifically, each proposed configuration is evaluated under explicit resource and time limits. Configurations that fail due to runtime errors, latency spikes, or resource exhaustion are assigned a zero reward, ensuring that the global search process remains well-defined. The environment additionally supports both corpus-level reward aggregation, suitable for standard black-box optimizers, and finer-grained feedback appropriate for reinforcement-learning methods.

\subsection{Algorithm Details}
\label{app:controller_details}

This appendix provides additional details for the 13 algorithms summarized in Table~\ref{tab:algo_taxonomy_13}.

\subsubsection{Sampling and Local Search}

\paragraph{Random Search.}
Random Search \cite{bergstra2012random} samples complete configurations independently from the discrete search space, providing the reference baseline for all structured optimizers.

\paragraph{Greedy Search.}
Greedy Search starts from a current configuration and applies the locally best modification available at each step. It quantifies how far the search space can be navigated by pure exploitation.

\paragraph{Coordinate Descent.}
Coordinate Descent \cite{wright2015coordinate} optimizes one hyperparameter dimension while holding the others fixed, and is most effective when the objective is partially separable across modules.

\paragraph{Simulated Annealing.}
Simulated Annealing \cite{kirkpatrick1983annealing} performs local moves but accepts lower-scoring configurations with a temperature-controlled probability, allowing the search to escape poor local optima early and become more selective later.

\paragraph{Iterative Local Search.}
Iterative Local Search \cite{lourenco2003iterated} alternates between local improvement and explicit perturbation. Relative to single-trajectory local search, it reduces dependence on initialization.

\subsubsection{SMBO, Distribution-Guided, and Evolutionary Search}

\paragraph{Tree-structured Parzen Estimator (TPE).}
TPE is an SMBO method \cite{bergstra2011algorithms} related to Bayesian optimization \cite{snoek2012practical}. It partitions observed configurations into high- and low-reward sets and fits separate density models to the two regions; new candidates are then selected to favor configurations with high expected improvement under this surrogate.

\paragraph{Cross-Entropy Method.}
The Cross-Entropy Method \cite{rubinstein1999crossentropy} is treated here as an estimation-of-distribution-style optimizer: it maintains a parameterized sampling distribution over configurations and updates it from the current elite set. In discrete spaces, this yields a direct model of which choices should accumulate more probability mass.

\paragraph{Regularized Evolution.}
Regularized Evolution \cite{real2019regularized} maintains a finite population, samples parents from it, and generates offspring through mutation. Aging-based regularization removes old individuals and prevents premature population collapse.

\subsubsection{Bandit Search}

\paragraph{Thompson Sampling.}
Thompson Sampling \cite{thompson1933likelihood} chooses actions by sampling from posterior reward estimates, yielding stochastic exploration proportional to uncertainty. In our setting, it serves as a bandit-style baseline for adaptive selection.

\paragraph{Upper Confidence Bound (UCB).}
UCB \cite{auer2002finite} selects the action with the largest optimistic score
\begin{equation}
    \text{Score}_{\mathrm{UCB}} = \bar{r}_i + c \sqrt{\frac{\ln N}{n_i}},
\end{equation}
where $\bar{r}_i$ is the empirical mean reward of action $i$, $n_i$ its pull count, and $N$ the total number of decisions. The rule renders the exploration term explicit and deterministic.

\subsubsection{RL-based Methods}

\paragraph{GRPO.}
GRPO \cite{shao2024deepseekmath} treats configuration generation as a policy over discrete choices and updates the policy with group-relative advantages. This reduces variance relative to raw score-based updates and enables direct policy learning from black-box rewards.

\paragraph{Dr.~GRPO.}
Dr.~GRPO \cite{liu2025drgrpo} is a GRPO variant designed to reduce optimization bias and improve token efficiency. In our benchmark, we use it as a conservative GRPO-style baseline for noisy, small-budget settings.

\paragraph{Reinforce++.}
Reinforce++ \cite{hu2025reinforcepp} optimizes a stochastic policy with critic-free, globally normalized policy-gradient updates. It serves as a lighter RL baseline against which the more specialized GRPO variants can be compared, while remaining closely related to classical REINFORCE-style updates \cite{williams1992reinforce}.

\subsection{Benchmark Protocol Details}
\label{app:benchmark_protocol_details}

The main algorithm comparison is conducted on the seven benchmark tasks listed in Table~\ref{tab:datasets}. Each algorithm is evaluated under the corresponding instantiated text or multimodal search space, with seeds $11$, $22$, and $33$ and a uniform budget of 30 configuration evaluations per seed. This protocol isolates search behavior from changes in pipeline structure, dataset size, or metric definition, while reducing dependence on a single random seed.

The scalar objective is an equal-weight aggregate of ROUGE-L, METEOR, token-F1, and BLEU. In LLM settings, rewriting, pruning, and generation are performed by the Qwen3 text stack. All algorithms interact with the same black-box \texttt{evaluate(config)} interface and observe the same proxy dataset for each task.

This shared protocol is intended to keep the main comparison focused on search behavior rather than on metric selection or implementation differences. It thus enables controlled comparison across algorithm families while holding the end-to-end RAG environment fixed.

These settings are intentionally modest: the proxy benchmark keeps evaluation fast enough for repeated optimization while still exposing differences between controllers, and 30 evaluations per seed provides a fixed budget for comparing search efficiency. Using three seeds helps separate algorithmic effects from sampling noise, while the shared \texttt{evaluate(config)} interface ensures that all methods face the same pipeline implementation and scoring rule. Together, these choices make the benchmark suitable for controlled comparison rather than for absolute system tuning.

\subsection{Reproducibility Details}
\label{app:reproducibility_details}

Each benchmark environment is stored as a pair of files. The QA file contains query and reference fields, while the corpus file contains the retrieval units used by the pipeline. For LongBench-Multifield, LongBench-Qasper, and ScienceQA, the proxy subsets are sampled with a fixed subset seed of 42. ScienceQA additionally keeps examples with a valid answer, a non-empty hint, and an available image, and exports the image path into the corpus entry. These fixed proxy files are used unchanged across all algorithms and seeds.

All text generation calls use the same local LLM configuration: temperature 0, maximum output length 256 tokens, and a 60-second request timeout. LLM runs use Qwen3-4B-Instruct for rewriting, pruning, and generation, while the multimodal LLM run uses Qwen3-VL-4B-Instruct for generation with CLIP-based visual retrieval. Experiments were run on a two-GPU workstation with PRO6000 GPUs.

A configuration evaluation means running one complete pipeline configuration over the full proxy environment and scoring the resulting answers with the selected objective. The evaluation cache is keyed by the configuration, dataset file hashes, modality, and evaluation mode, so repeated configurations return the same cached reward. Failed configurations receive zero reward and remain part of the budgeted trial sequence.

\section{Detailed Results}

\subsection{Detailed Main Results}
\label{app:main_results_details}

Tables~\ref{tab:hotpot_detailed_results}--\ref{tab:scienceqa_detailed_results} report dataset-level algorithm results for the main benchmark. Each algorithm is evaluated with a budget of 30 trials under seeds $11$, $22$, and $33$. The ranking score is the equal-weight aggregate of ROUGE-L, METEOR, token-F1, and BLEU. The tables report each seed separately and summarize the mean and standard deviation across seeds.

For compactness, we adopt the following abbreviations: Coord. = Coordinate Descent, SA = Simulated Annealing, ILS = Iterative Local Search, CEM = Cross-Entropy Method, Reg-Evo = Regularized Evolution, TS = Thompson Sampling, and Dr. GRPO = Dr. GRPO.

\begin{table*}[t]
\centering
\footnotesize
\setlength{\tabcolsep}{4pt}
\renewcommand{\arraystretch}{1.0}
\begin{tabular*}{\textwidth}{@{\extracolsep{\fill}}lccccc@{}}
\toprule
Alg. & Seed 11 & Seed 22 & Seed 33 & Mean & Std. \\
\midrule
Greedy & 0.4109 & 0.4205 & 0.4207 & 0.4174 & 0.0046 \\
Coord. & 0.3798 & 0.4205 & 0.4255 & 0.4086 & 0.0205 \\
Reg-Evo & 0.4064 & 0.3746 & 0.4217 & 0.4009 & 0.0196 \\
Dr. GRPO & 0.4071 & 0.4219 & 0.3121 & 0.3803 & 0.0487 \\
Reinforce++ & 0.4035 & 0.3430 & 0.3701 & 0.3722 & 0.0247 \\
SA & 0.3254 & 0.3605 & 0.4171 & 0.3677 & 0.0378 \\
TPE & 0.3998 & 0.2568 & 0.4193 & 0.3586 & 0.0724 \\
Random & 0.3252 & 0.4145 & 0.3208 & 0.3535 & 0.0432 \\
TS & 0.3828 & 0.2992 & 0.3687 & 0.3502 & 0.0366 \\
ILS & 0.3777 & 0.3996 & 0.2417 & 0.3397 & 0.0698 \\
GRPO & 0.3968 & 0.3442 & 0.2587 & 0.3332 & 0.0569 \\
CEM & 0.3140 & 0.3114 & 0.3425 & 0.3226 & 0.0141 \\
UCB & 0.2759 & 0.2479 & 0.3969 & 0.3069 & 0.0647 \\
\bottomrule
\end{tabular*}
\caption{Three-seed detailed results for HotpotQA. Scores are best-of-30 weighted objectives.}
\label{tab:hotpot_detailed_results}
\end{table*}

\begin{table*}[t]
\centering
\footnotesize
\setlength{\tabcolsep}{4pt}
\renewcommand{\arraystretch}{1.0}
\begin{tabular*}{\textwidth}{@{\extracolsep{\fill}}lccccc@{}}
\toprule
Alg. & Seed 11 & Seed 22 & Seed 33 & Mean & Std. \\
\midrule
CEM & 0.2451 & 0.2248 & 0.2469 & 0.2389 & 0.0100 \\
Reinforce++ & 0.2483 & 0.2350 & 0.2247 & 0.2360 & 0.0097 \\
Random & 0.2465 & 0.2366 & 0.2129 & 0.2320 & 0.0141 \\
Coord. & 0.2255 & 0.2791 & 0.1882 & 0.2309 & 0.0373 \\
TPE & 0.2204 & 0.2241 & 0.2172 & 0.2206 & 0.0028 \\
Dr. GRPO & 0.2483 & 0.2310 & 0.1768 & 0.2187 & 0.0305 \\
Reg-Evo & 0.2224 & 0.2242 & 0.2082 & 0.2183 & 0.0072 \\
TS & 0.2220 & 0.2241 & 0.1970 & 0.2144 & 0.0123 \\
GRPO & 0.1774 & 0.2219 & 0.2417 & 0.2137 & 0.0269 \\
ILS & 0.2100 & 0.2127 & 0.1974 & 0.2067 & 0.0067 \\
SA & 0.2101 & 0.1829 & 0.2226 & 0.2052 & 0.0166 \\
UCB & 0.2181 & 0.1729 & 0.2236 & 0.2049 & 0.0227 \\
Greedy & 0.1910 & 0.2179 & 0.1882 & 0.1990 & 0.0134 \\
\bottomrule
\end{tabular*}
\caption{Three-seed detailed results for MS MARCO. Scores are best-of-30 weighted objectives.}
\label{tab:msmarco_detailed_results}
\end{table*}

\begin{table*}[t]
\centering
\footnotesize
\setlength{\tabcolsep}{4pt}
\renewcommand{\arraystretch}{1.0}
\begin{tabular*}{\textwidth}{@{\extracolsep{\fill}}lccccc@{}}
\toprule
Alg. & Seed 11 & Seed 22 & Seed 33 & Mean & Std. \\
\midrule
GRPO & 0.2372 & 0.2288 & 0.2268 & 0.2309 & 0.0045 \\
TPE & 0.2297 & 0.2199 & 0.2280 & 0.2258 & 0.0043 \\
CEM & 0.2256 & 0.2225 & 0.2269 & 0.2250 & 0.0018 \\
Reg-Evo & 0.2209 & 0.2256 & 0.2265 & 0.2243 & 0.0025 \\
Coord. & 0.2231 & 0.2288 & 0.2192 & 0.2237 & 0.0040 \\
Greedy & 0.2217 & 0.2279 & 0.2191 & 0.2229 & 0.0037 \\
Random & 0.2160 & 0.2171 & 0.2356 & 0.2229 & 0.0090 \\
TS & 0.2079 & 0.2199 & 0.2226 & 0.2168 & 0.0064 \\
Reinforce++ & 0.1939 & 0.2156 & 0.2273 & 0.2123 & 0.0139 \\
ILS & 0.2252 & 0.1987 & 0.1866 & 0.2035 & 0.0161 \\
UCB & 0.1916 & 0.1956 & 0.2196 & 0.2023 & 0.0124 \\
Dr. GRPO & 0.2126 & 0.1957 & 0.1966 & 0.2017 & 0.0078 \\
SA & 0.2006 & 0.1789 & 0.2195 & 0.1996 & 0.0166 \\
\bottomrule
\end{tabular*}
\caption{Three-seed detailed results for SQuADv2. Scores are best-of-30 weighted objectives.}
\label{tab:squadv2_detailed_results}
\end{table*}

\begin{table*}[t]
\centering
\footnotesize
\setlength{\tabcolsep}{4pt}
\renewcommand{\arraystretch}{1.0}
\begin{tabular*}{\textwidth}{@{\extracolsep{\fill}}lccccc@{}}
\toprule
Alg. & Seed 11 & Seed 22 & Seed 33 & Mean & Std. \\
\midrule
Greedy & 0.4195 & 0.3820 & 0.4104 & 0.4040 & 0.0160 \\
SA & 0.3949 & 0.4151 & 0.3969 & 0.4023 & 0.0091 \\
GRPO & 0.4206 & 0.4054 & 0.3792 & 0.4017 & 0.0171 \\
Random & 0.4072 & 0.4037 & 0.3934 & 0.4015 & 0.0059 \\
Coord. & 0.4030 & 0.3665 & 0.4214 & 0.3970 & 0.0228 \\
Dr. GRPO & 0.3917 & 0.3994 & 0.3951 & 0.3954 & 0.0032 \\
Reg-Evo & 0.3872 & 0.3879 & 0.4059 & 0.3937 & 0.0087 \\
Reinforce++ & 0.4018 & 0.3812 & 0.3975 & 0.3935 & 0.0089 \\
CEM & 0.3648 & 0.4163 & 0.3893 & 0.3901 & 0.0210 \\
UCB & 0.4018 & 0.3648 & 0.4006 & 0.3891 & 0.0171 \\
TPE & 0.3525 & 0.3886 & 0.4185 & 0.3865 & 0.0270 \\
TS & 0.3525 & 0.3727 & 0.4020 & 0.3757 & 0.0203 \\
ILS & 0.4064 & 0.3492 & 0.3714 & 0.3756 & 0.0236 \\
\bottomrule
\end{tabular*}
\caption{Three-seed detailed results for TriviaQA. Scores are best-of-30 weighted objectives.}
\label{tab:triviaqa_detailed_results}
\end{table*}

\begin{table*}[t]
\centering
\footnotesize
\setlength{\tabcolsep}{4pt}
\renewcommand{\arraystretch}{1.0}
\begin{tabular*}{\textwidth}{@{\extracolsep{\fill}}lccccc@{}}
\toprule
Alg. & Seed 11 & Seed 22 & Seed 33 & Mean & Std. \\
\midrule
Reg-Evo & 0.0990 & 0.1092 & 0.0971 & 0.1018 & 0.0053 \\
TPE & 0.0935 & 0.1098 & 0.0989 & 0.1007 & 0.0068 \\
Dr. GRPO & 0.0870 & 0.1009 & 0.0964 & 0.0948 & 0.0058 \\
Reinforce++ & 0.0944 & 0.1010 & 0.0883 & 0.0946 & 0.0052 \\
Coord. & 0.0929 & 0.0930 & 0.0976 & 0.0945 & 0.0022 \\
Random & 0.0985 & 0.0917 & 0.0873 & 0.0925 & 0.0046 \\
SA & 0.0963 & 0.0974 & 0.0826 & 0.0921 & 0.0067 \\
Greedy & 0.0926 & 0.0849 & 0.0976 & 0.0917 & 0.0052 \\
CEM & 0.1019 & 0.0866 & 0.0828 & 0.0904 & 0.0082 \\
TS & 0.0829 & 0.0874 & 0.0980 & 0.0894 & 0.0063 \\
UCB & 0.0784 & 0.0901 & 0.0973 & 0.0886 & 0.0078 \\
GRPO & 0.0745 & 0.0918 & 0.0918 & 0.0861 & 0.0082 \\
ILS & 0.0956 & 0.0730 & 0.0824 & 0.0837 & 0.0092 \\
\bottomrule
\end{tabular*}
\caption{Three-seed detailed results for LongBench-Qasper. Scores are best-of-30 weighted objectives.}
\label{tab:qasper_detailed_results}
\end{table*}

\begin{table*}[t]
\centering
\footnotesize
\setlength{\tabcolsep}{4pt}
\renewcommand{\arraystretch}{1.0}
\begin{tabular*}{\textwidth}{@{\extracolsep{\fill}}lccccc@{}}
\toprule
Alg. & Seed 11 & Seed 22 & Seed 33 & Mean & Std. \\
\midrule
SA & 0.3385 & 0.3883 & 0.3597 & 0.3622 & 0.0204 \\
TPE & 0.3600 & 0.3641 & 0.3621 & 0.3621 & 0.0017 \\
Reg-Evo & 0.3465 & 0.3613 & 0.3701 & 0.3593 & 0.0097 \\
GRPO & 0.3579 & 0.3552 & 0.3331 & 0.3487 & 0.0111 \\
CEM & 0.3316 & 0.3511 & 0.3616 & 0.3481 & 0.0124 \\
Dr. GRPO & 0.3570 & 0.3295 & 0.3565 & 0.3477 & 0.0129 \\
TS & 0.3314 & 0.3429 & 0.3621 & 0.3455 & 0.0127 \\
Greedy & 0.3729 & 0.2760 & 0.3677 & 0.3389 & 0.0445 \\
Coord. & 0.3576 & 0.2857 & 0.3677 & 0.3370 & 0.0365 \\
Random & 0.3522 & 0.3030 & 0.3535 & 0.3363 & 0.0235 \\
UCB & 0.3314 & 0.2958 & 0.3738 & 0.3337 & 0.0319 \\
ILS & 0.3634 & 0.2944 & 0.3358 & 0.3312 & 0.0284 \\
Reinforce++ & 0.3092 & 0.3365 & 0.3380 & 0.3279 & 0.0133 \\
\bottomrule
\end{tabular*}
\caption{Three-seed detailed results for LongBench-Multifield. Scores are best-of-30 weighted objectives.}
\label{tab:multifield_detailed_results}
\end{table*}

\begin{table*}[t]
\centering
\footnotesize
\setlength{\tabcolsep}{4pt}
\renewcommand{\arraystretch}{1.0}
\begin{tabular*}{\textwidth}{@{\extracolsep{\fill}}lccccc@{}}
\toprule
Alg. & Seed 11 & Seed 22 & Seed 33 & Mean & Std. \\
\midrule
Random & 0.3244 & 0.3101 & 0.3101 & 0.3149 & 0.0067 \\
Coord. & 0.3180 & 0.3101 & 0.3101 & 0.3127 & 0.0037 \\
Greedy & 0.3182 & 0.3101 & 0.3046 & 0.3110 & 0.0056 \\
GRPO & 0.2451 & 0.3101 & 0.3101 & 0.2884 & 0.0307 \\
Dr. GRPO & 0.2063 & 0.3102 & 0.3101 & 0.2755 & 0.0490 \\
ILS & 0.1935 & 0.3044 & 0.3098 & 0.2693 & 0.0536 \\
TPE & 0.2515 & 0.3099 & 0.2461 & 0.2692 & 0.0289 \\
SA & 0.1784 & 0.3099 & 0.3101 & 0.2661 & 0.0620 \\
Reg-Evo & 0.2536 & 0.2280 & 0.3101 & 0.2639 & 0.0343 \\
TS & 0.3242 & 0.2217 & 0.2197 & 0.2552 & 0.0488 \\
Reinforce++ & 0.1371 & 0.3102 & 0.3101 & 0.2525 & 0.0816 \\
UCB & 0.3102 & 0.2181 & 0.2152 & 0.2478 & 0.0441 \\
CEM & 0.3188 & 0.1346 & 0.2371 & 0.2302 & 0.0753 \\
\bottomrule
\end{tabular*}
\caption{Three-seed detailed results for ScienceQA. Scores are best-of-30 weighted objectives.}
\label{tab:scienceqa_detailed_results}
\end{table*}

\subsection{Random-Average Ablation Details}
\label{app:random_average_ablation_details}

Table~\ref{tab:random_average_ablation_details} reports the full per-algorithm results for the HotpotQA random-average ablation. All runs use a single seed and a budget of 20 evaluations. The random-trial mean baseline is 0.0660, and the Random Search best-of-budget score is 0.1134; the latter is included only as a reference point and is not the ablation baseline. The $\Delta$ column is always computed against the random-trial mean. BERTScore-F1 uses the corrected evaluation values.

\begin{table*}[t]
\centering
\footnotesize
\setlength{\tabcolsep}{3pt}
\renewcommand{\arraystretch}{1.0}
\begin{tabular*}{\textwidth}{@{\extracolsep{\fill}}lcccccccc@{}}
\toprule
Alg. & Score & $\Delta$ & BERT & BLEU & F1 & Judge & MET. & R-L \\
\midrule
Dr. GRPO & 0.1247 & +0.0588 & 0.4388 & 0.0364 & 0.1475 & 0.5400 & 0.1712 & 0.1439 \\
Reinforce++ & 0.1247 & +0.0588 & 0.4388 & 0.0364 & 0.1475 & 0.5400 & 0.1712 & 0.1439 \\
UCB & 0.1211 & +0.0551 & 0.4463 & 0.0371 & 0.1438 & 0.5300 & 0.1611 & 0.1425 \\
Greedy & 0.1189 & +0.0529 & 0.4401 & 0.0368 & 0.1419 & 0.4800 & 0.1580 & 0.1387 \\
GRPO & 0.1180 & +0.0520 & 0.4379 & 0.0330 & 0.1367 & 0.5500 & 0.1695 & 0.1329 \\
Coord. & 0.1146 & +0.0487 & 0.4361 & 0.0292 & 0.1338 & 0.4800 & 0.1637 & 0.1317 \\
Random & 0.1134 & +0.0474 & 0.4390 & 0.0305 & 0.1362 & 0.5400 & 0.1542 & 0.1325 \\
Reg-Evo & 0.1103 & +0.0443 & 0.4360 & 0.0303 & 0.1313 & 0.5400 & 0.1503 & 0.1293 \\
SA & 0.1072 & +0.0413 & 0.4369 & 0.0263 & 0.1265 & 0.5200 & 0.1521 & 0.1240 \\
CEM & 0.1034 & +0.0375 & 0.4160 & 0.0319 & 0.1184 & 0.4700 & 0.1458 & 0.1176 \\
TPE & 0.0976 & +0.0316 & 0.4175 & 0.0317 & 0.1096 & 0.4500 & 0.1406 & 0.1084 \\
TS & 0.0930 & +0.0271 & 0.4083 & 0.0289 & 0.1034 & 0.4900 & 0.1384 & 0.1013 \\
ILS & 0.0437 & -0.0222 & 0.3404 & 0.0046 & 0.0407 & 0.3600 & 0.0918 & 0.0378 \\
\bottomrule
\end{tabular*}
\caption{Full results for the random-average ablation on HotpotQA.}
\label{tab:random_average_ablation_details}
\end{table*}

\end{document}